%% file: main.tex
\documentclass[10pt,twocolumn,letterpaper]{article}

\usepackage{ijcb}
\usepackage{times}
\usepackage{epsfig}
\usepackage{graphicx}
\usepackage{amsmath}
\usepackage{amssymb}
\usepackage{hhline}
\usepackage{color, colortbl}
\usepackage{multirow}
\usepackage{times}
\usepackage{cancel}
\usepackage{epsfig}
\usepackage{graphicx}
\usepackage{amsmath}
\usepackage{amssymb}
\usepackage{subcaption}
\usepackage[utf8]{inputenc}
\usepackage{lipsum} 
\usepackage{booktabs}
\usepackage{hyperref}
\usepackage{makecell}
\usepackage{url}
\usepackage{algorithm}
\usepackage{algpseudocode}
\usepackage{multirow}
\usepackage[table,xcdraw]{xcolor}

\definecolor{lightgray}{gray}{0.9}
\definecolor{lightblue}{rgb}{0.93,0.95,1.0}
\definecolor{darkgreen}{rgb}{0.0,0.6,0.0}
\definecolor{blue}{rgb}{1, 0, 0}

\usepackage[symbol]{footmisc}

\hypersetup{
    colorlinks=true,
    linkcolor=blue,
    filecolor=magenta,      
    urlcolor=cyan,
    pdftitle={Overleaf Example},
    pdfpagemode=FullScreen,
}

\ijcbfinalcopy % *** Uncomment this line for the final submission
\ifijcbfinal\fi

\begin{document}

%%%%%%%%% TITLE
\title{
The Gender Gap in Face Recognition Accuracy Is a Hairy Problem
}

% uncomment for the final submission as it doesn't work with the anonymous version
\author{Aman Bhatta$^1$, Vitor Albiero$^1$, Kevin W. Bowyer$^1$, Michael C. King$^2$
\\
$^1$ University of Notre Dame, $^2$ Florida Institute of Technology}
% \author[1]{Aman Bhatta}
% \author[1]{Vítor Albiero}
% \author[1]{Kevin W. Bowyer}
% \author[2]{Michael C. King}
% \affil[1]{University of Notre Dame, Notre Dame, Indiana}
% \affil[2]{Florida Institute of Technology, Melbourne, Florida}

\maketitle
\thispagestyle{empty}

\begin{abstract}
It is broadly accepted that there is a ``gender gap’’ in face recognition accuracy, with females having higher false match and false non-match rates.
However, relatively little is known about the cause(s) of this gender gap.
Even the recent NIST report on demographic effects lists ``analyze cause and effect'' under ``what we did not do''. 
We first demonstrate that female and male hairstyles have important differences that impact face recognition accuracy.
In particular, compared to females, male facial hair contributes to creating a greater average difference in appearance between different male faces.
We then demonstrate that when the data used to estimate recognition accuracy is balanced across gender for how hairstyles occlude the face, the initially observed gender gap in accuracy largely disappears.
We show this result for two different matchers,
and analyzing images of Caucasians and of African-Americans.
These results suggest that future research on demographic variation in accuracy should include a check for balanced quality of the test data as part of the problem formulation.
To promote reproducible research, matchers, attribute classifiers, and datasets used in this research are/will be publicly available. 
\end{abstract}

\section{Introduction}
Deep learning algorithms rule the world of face recognition research.  
Thus the natural reaction to observing different face recognition accuracy across demographic groups is to point to imbalance in the quantity of training data across the demographic groups.
But what if the difference in test accuracy is {\it not} caused by imbalance in the quantity of training data?
What if the difference in test accuracy is caused by imbalance in the {\it quality} of {\it test} data?
We show that the observed gender gap in face recognition accuracy largely disappears when female and male test images are balanced on basic elements of how hairstyle affects appearance.

\input{latex_figures/intro_fig}

The observation that face recognition algorithms achieve different accuracy for females and males goes back at least to % the Face Recognition Vendor Test 
2002~\cite{Frvt_2002}.
The general observation that has been repeated in various recent research results, including the NIST report on demographic effects~\cite{Frvt_demographic}, is that females tend to have a worse impostor distribution (higher false-match rate) and worse genuine distribution (higher false non-match rate).
We confirm this effect %We begin by observing that this is true
for both African-Americans and for Caucasians, for both the ArcFace matcher {\it and also for a matcher trained with explicitly gender-balanced training data}. % (Figure~\ref{fig:original_dist}).
We then consider some basic elements of hairstyle that differ strongly between females and males,
% (Figure~\ref{fig:hair_ratio} and Table~\ref{tab:count}), 
and how those elements impact face recognition accuracy. 
% (Figure~\ref{fig:sep_attribute}).
Based on this, we select a fair subset of the original test data that has hairstyle-balanced female / male face visibility.
The originally-observed gender gap in face recognition accuracy largely disappears, for both matchers, when accuracy is evaluated with the gendered-hairstyle-neutral test data.
% (Figure~\ref{fig:exp_results} and Table~\ref{tab:Results_Table}).

This work suggests that differences in hairstyle-related occlusion of the face are sufficient to explain nearly all of the observed difference in face recognition accuracy between females and males.
To promote open and reproducible research, the face matchers, face attribute classifiers and datasets used in this research are all ones that are available to other researchers.

\section{Related Work}

Face recognition accuracy differences across demographic groups have been observed in research results at least since 2002~\cite{Frvt_2002}.
In recent years, the topic has attracted substantial attention from news media~\cite{Doctorow2019, Hoggins2019,Santow2020,Vincent2019}.
% and from the research community (cite a bunch).
For a broad overview of the topic, see recent surveys \cite{Abdurrahim2018,Drozdowski2020}.
Here we briefly touch on selected related works.

The earliest work we are aware of to report lower accuracy for females is the 2002 Face Recognition Vendor Test (FRVT)~\cite{Frvt_2002}. Evaluating ten algorithms of the pre-deep learning network era, identification rates of the top systems are 6\% to 9\% higher for males. Klare et al.~\cite{Klare1} analyzed demographic accuracy differences using multiple matchers (commercial, nontrainable, and trainable) and reported that “The female, Black, and younger cohorts have worse ROC curves and thus have lower accuracy''. 
They also showed impostor and genuine distributions with the same qualitative relation as shown in the results in Figure~\ref{fig:original_dist}.

There is relatively little work that attempts to identify the cause of the gender gap in face recognition accuracy. For example, even the most extensive study in this area, the recent NIST report~\cite{Frvt_demographic} on demographic effects, lists ``analyze cause and effect'' under the heading of ``what we did not do''. 

Past researchers have speculated causes such as the use of cosmetics~\cite{Cook1,Klare1,Lu1}, more varied hairstyles~\cite{Albiero_gender}, or shorter height for women, leading to non-optimal camera angle~\cite{Cook1,Grother2011}. Since the advent of deep learning, imbalanced training data is often suggested as the cause \cite{Drozdowski2020,Merler2019,Vera1}.
% (cite a bunch here). 
Few works have made any experimental analysis to attempt to determine the cause or causes.

Imbalance in the training data is explored as a possible cause of the
the gap in recognition accuracy in~\cite{Albiero_balance}. Experiments with VGGFace2~\cite{Vggface2} and MS-Celeb~\cite{MS1}, various loss functions, and multiple test sets did not reveal evidence supporting the premise that gender-balanced training data results in balanced accuracy on test data. Experiments examining different pose, expression, makeup use and forehead occlusion by hair between females and males in the MORPH dataset were reported in~\cite{Albiero_gender}.  Differences were found between females and males in each of these
factors, but balancing for the factors individually or together did not equalize female/male accuracy in the test data.

Research on how facial hair and hairstyles affect recognition accuracy is limited. Studies published before the deep-learning era~\cite{Givens1,Givens2} reported that accuracy is better if there is facial hair in one of the images. However, Lu et al.~\cite{Lu1} used deep learning matchers to study effects of facial hair and reported that facial hair does not change the key features of faces, and state-of-the-art deep learning models can handle most facial hair variations. Terhörst et al.~\cite{Terhorst_2022} investigate the influence of 47 attributes on the verification accuracy using the MAADface datset. 
However, this work does not evaluate effects of hair-related attributes across demographics.
Also, the MAADFace dataset is based on VGGFace2 \cite{Vggface2}, and issues have been reported with the accuracy of identity labels in VGGFace2~\cite{albiero2019does}, which may carry over to affect MAADFace.

The closest related works are %most closely related previous work is
~\cite{Albiero_BMVC, AlbieroTIFS}, which attempt to explain the difference in female/male genuine distributions by balancing the test image sets on the fraction of the image occupied by the face. However, they do not take into account facial hair and beards.  Also, they do not clearly identify a cause for the difference in female/male impostor distributions, and speculate that it is due to biological differences in face appearance. Our work goes beyond this by using a more detailed and complete analysis of hair and hairstyle to show that the observed female/male differences in both impostor and genuine distributions can be largely accounted for by differences in hairstyle.

\input{latex_figures/original_dists}

\section{Baseline Gender Gap In Accuracy} 

The impostor and genuine distributions in Figure~\ref{fig:original_dist} are representative of the gender gap in recognition accuracy observed by various researchers.  The top row shows results from ArcFace and the bottom row shows results from a matcher whose training data was explicitly balanced on number of female and male identities and images~\cite{Albiero_balance}. 
The plots in the first column of Figure~\ref{fig:original_dist} are based on the Caucasian subset of the MORPH dataset~\cite{morph_site, morph}, the plots in the second column are for the African-American subset of MORPH, and the last column is based on a smaller dataset of our own, comprised of Caucasian images. 
In all six instances of (different matcher $\times$ different racial group), the female impostor distribution (and so the FMR) and the female genuine distribution (and so the FNMR) are worse. 
This is the baseline ``gender gap'' that this work seeks to understand the cause of. To quantify the gap, the d-prime difference between the corresponding female and male distribution is given. A larger d-prime indicates a larger gap between the female and male distributions.

The MORPH dataset~\cite{morph_site, morph} was initially collected to support research in face aging and it has recently been widely used in the study of demographic variation in accuracy~\cite{Albiero_BMVC, Albiero_gender, Albiero_balance,Drozdowski2021, Georgopoulos2021, KrishnapriyaTTS, Vangara1}.
MORPH contains mugshot-style images that are approximately frontal pose, neutral expression, acquired with controlled lighting, and have an 18\% gray background.  
% Several increasing-size versions of MORPH have been released. For 
In this work, we use the same version of MORPH that \cite{AlbieroTIFS} reported results on. This version has 35,276 images of 8,835 Caucasian males, 10,941 images of 2,798 Caucasian females, 56,245 images of 8,839 African-American males, and 24,857 images of 5,929 African-American females.
MORPH faces were detected and aligned using~\cite{albiero2021img2pose}.
MORPH is particularly appropriate for this research, as noted by Drozdowski et al.~\cite{Drozdowski2020}, ``due to its large size, relatively constrained image acquisition conditions, and the presence of ground-truth labels (from public records) for sex, race, and age of the subjects''.

The second dataset used is one acquired, with human subjects approval, at our own institution (to be released after publication).
Using a second dataset acquired independently of MORPH guards against results being an accident of a particular dataset.
Images in our dataset were acquired indoors, and are roughly frontal pose and neutral expression. The dataset is composed of 5,444 images of 575 Caucasian females and 7,003 images of 687 Caucasian males. 

ArcFace is a popular deep-learning face matcher with state-of-the-art accuracy~\cite{Arcface}.
ArcFace uses the Additive Angular Margin Loss function to optimize the feature embedding to enforce higher similarity for intra-class samples and diversity for inter-class samples. The instance of ArcFace used here is publicly-available~\cite{Insightface}.
%, trained on the MS1MV2 dataset~\cite{MS1}. 
The input to ArcFace is an aligned face resized to 112x112, and the output is a 512-d feature vector that is matched using cosine similarity.
The gender-balanced matcher~\cite{Albiero_balance} used here is an ArcFace-like, ResNet-based~\cite{resnet} matcher whose training data is explicitly balanced on number of female and male identities and images, and we use the publicly-available trained model~\cite{Albiero_balance}.
% It was trained using a combined margin loss composed of ArcFace~\cite{Arcface}, CosFace~\cite{Cosface} and SphereFace~\cite{Sphereface} margins, 
Again, the input is a 112x112 face image, and the output is a 512-d feature vector that is matched using cosine similarity.
We will refer to this matcher as gender-balanced matcher.
% As ArcFace is the most known open-source state-of-the-art (SoTA) matcher, it was chosen as a representative of other open-source SoTA matchers. %The point of using ArcFace is to have a well-known state-of-the-art reference point for accuracy.
The gender-balanced matcher is used here to have results from a second matcher, and to emphasize that balancing number of identities and images in the training data is no guarantee of balanced accuracy on test data. 

\input{latex_tables/count}

\section{Gender-Based Differences In Hairstyle}

This section discusses three dimensions of hairstyle: baldness; beard, mustache and other facial hair;  %– (1) baldness, (2) beard, mustache and other facial hair, and (3) 
the ``size'' of the hairstyle as measured by the fraction of the 112x112 cropped face image that is occupied by hair.
Each of these three dimensions of hair is observed to have large differences between female and male face images.
And a difference in any of these three dimensions can cause a noticeable difference in the impostor and / or genuine distribution.

\subsection{Bald Hairstyle}

A bald hairstyle is simply one in which there is little to no visible hair on the top of the head. A person may choose to have a bald hairstyle, or may be bald due to age or medical condition. On average, a person who is bald and another who is not bald look less alike, compared to two persons who are both bald, or to two who are both not bald.

We used a fusion of the modified Bilateral Segmentation network (``BiSeNet'') algorithm~\cite{Bisenet_github, Bisenet} results and Microsoft Face API~\cite{Microsoft_api} results to automatically detect face images with a bald hairstyle. A pre-trained version of BiSeNet segments a face image into semantic regions, with region 17 corresponding to hair flowing from the top of the head, not including facial hair such as beard or mustache.
A 112x112 cropped (frontal) face image with less than 2\% of its pixels labeled as hair by BiSeNet generally corresponds to a bald hairstyle.
The Microsoft Face API predicts baldness with a confidence ranging from 0 to 1.
We found that a threshold of 0.97 results in high confidence for a bald hairstyle.
We use a simple conjunction to fuse the results of the two algorithms, and label an image as having a bald hairstyle if (a) less than 2\% of pixels are labeled as hair in the BiSeNet segmentation and (b) the Microsoft Face API baldness prediction is greater than or equal to 0.97.

Using this fusion algorithm, the fraction of MORPH and our own data labeled as bald hairstyle for each of the demographic groups is given in Table~\ref{tab:count}.
For MORPH, almost no female images were labeled as having bald hairstyle, just 0.1\% of Caucasian female and 0.2\% of African-American female.In contrast, 4\% of Caucasian male and over 10\% of African-American male images were labeled as having bald hairstyle. For our own dataset, no female images and 0.4\% of male images were labeled as having bald hairstyle.

As an example of how the frequency of bald hairstyle in a set of images can impact the observed face recognition accuracy, Figure~\ref{fig:sep_bald} shows the impostor and genuine differences for MORPH African-American male face images broken out by pairs of images that both have bald hairstyle, that are both not bald, and that are a bald/not-bald mix.  The impostor distribution for the bald/not-bald image pairs shows the lowest similarity, followed by impostor distribution for not-bald image pairs, and then bald pairs. 
Thus, on average, images of two different persons, one with bald hairstyle and one not, look less similar than images of two different persons who are both bald or both not bald.
\subsection{Beard and Facial Hair} \label{beard}
Beard, mustache, sideburns, five o’clock shadow and related facial hair are elements of hairstyle that are generally limited to male images.
There is great variety in the facial hair that males may choose as part of their hairstyle, including the option of clean-shaven (no facial hair).
On average, one male face image that is clean-shaven and one that has facial hair will appear less similar that two images that are both clean-shaven or that both have similar facial hair.

We used a three-part fusion of results from the Microsoft Face API and Amazon Rekognition~\cite{Amazon_api} to classify images as clean-shaven or having facial hair. Microsoft Face predicts presence of beard, mustache, and sideburns, individually, each with a confidence score that can take on values 0, 0.1, 0.4, 0.6 and 0.9.  In our experience, a confidence score of 0.6 or 0.9 is generally accurate for the presence of facial hair, but some instances of facial hair still occur at lower confidence values.  For this reason, images with Microsoft Face confidence less than 0.6 are filtered with results from Amazon Rekognition. Amazon Rekognition gives a True/False for facial hair along with a confidence score from 50 to 100. An Amazon Rekognition result of True with a confidence greater than 85 is taken as indicating facial hair.  Lastly, an image with a Microsoft Face confidence of 0.4 and an Amazon Rekognition True with confidence greater than 55 or an Amazon Rekognition False with confidence less than 65  is taken as indicating facial hair.  This fusion approach is reasonably accurate but not perfect in separating facial hair and clean-shaven images.

Using this fusion algorithm, the fraction of the MORPH and our own images labeled as facial hair / clean-shaven for each of the four demographic groups is given in Table~\ref{tab:count}.
For MORPH, almost no female images were labeled as having facial hair, but 90\% of African-American male and 70\% of Caucasian male images were labeled as having facial hair. For our own dataset, no female face images were labeled as having facial hair, but 25\% of Caucasian male images were labeled as having facial hair.

As an example of how the frequency of facial hair can impact recognition accuracy, Figure~\ref{fig:sep_beard} shows the impostor and genuine differences for MORPH Caucasian male face images broken out by pairs where one is classified as facial hair and one clean-shaven, both classified as facial hair, and both classified as clean-shaven.  Image pairs with one classified as facial hair and one clean-shaven have an impostor distribution and a genuine distribution with lower average similarity than pairs with both images having facial hair, or with both being clean-shaven.
\input{latex_figures/one_attribute}
\input{latex_figures/hair_ratio_display}

\subsubsection{Misclassification in Facial Hair Prediction}
As explained above, we use a fusion algorithm based on Microsoft Face and Amazon Rekognition results to classify a given image as having facial hair or not. 
To check the sensitivity of this algorithm, we randomly selected 300 images for MORPH African-American male and 300 for MORPH Caucasian male.
Each group of 300 had 100 with prominent beard and facial hair, 100 with a small amount of facial hair, and 100 appearing clean-shaven.
For Caucasian male, all 100 of the prominent facial hair group were classified as having facial hair, 97 of 100 with smaller facial hair were classified as having facial hair, and 81 of 100 with no facial hair were classified as clean-shaven/no-facial hair.
For African-American male, all 100 of the prominent facial hair group were classified as having facial hair, 99 of 100 with smaller facial hair were classified as having facial hair, and only 32 of 100 with no facial hair were classified as clean-shaven/no facial-hair.
This experiment shows that we can be fairly confident about the no-facial hair subsets used in our experiments.
However, it also shows that a substantial number of clean-shaven images will also be classified as having facial hair, and that the over-classification of clean-shaven images as having facial hair is worse for African-American.
This over-classification also shows in the number of African-American female images incorrectly classified as having facial hair as shown in Table \ref{tab:count}.
Small amounts of shadow or variation in skin pigmentation can cause a clean-shaven image to be incorrectly classified as having facial hair.
Developing a more accurate facial attribute classifier for facial hair would allow more accurate hairstyle-balanced comparisons of recognition accuracy.

\subsection{Fullness of Hairstyle}
\input{latex_figures/hair_ratio_dist}
\input{latex_figures/exp_results}

In general, an increasing fraction of the image containing hair means increasing occlusion of the face by hair, as illustrated in Figure~\ref{fig:hair_ratio_display}.  The distribution of the fraction of the image that is labeled as hair in the BiSeNet segmentation (the ``hair ratio''), is shown for the demographic groups in Figure~\ref{fig:hair_ratio}.  Note that for MORPH, both African-American and Caucasian males have a spike at 0\%, representing bald hairstyles, and then another broader peak under 20\%, representing hairstyles that generally do not occlude much of the face.  In contrast, Caucasian females have a broad peak in the 40\% to 50\% range, implying substantially more occlusion of the face by hair.  And African-American females have a broad plateau in the 10\% to 50\% range, indicating a varied range of hairstyles that occlude different amounts of the face.  For our own dataset, there is a peak for males at slightly above 20\%, whereas for women there is peak in the range 35\% to 45\%.
It is clear from the distributions in Figure~\ref{fig:hair_ratio} that females have a broader range of hairstyles than males, and that on average a female face image has more of the face occluded by hair than a male face image.

As an example of how different distributions of ``hair ratio'' translate into occlusion that impacts accuracy, we divide the MORPH Caucasian female distribution in Figure~\ref{fig:sep_hair} into a lower tail (below 25\% hair ratio) and an upper tail (above 50\% hair ratio).
 Figure~\ref{fig:sep_hair} shows the impostor and genuine distributions computed for image pairs in the lower tail (less occlusion of face by hair), the upper tail (more occlusion of face by hair), and across the lower and upper tail (different patterns of hair occluding face).
Image pairs from across the tails result in an impostor distribution and a genuine distribution centered at lower similarity than the distributions from image pairs with similar face occlusion by hair.
Image pairs from either tail result in similar impostor distributions, but image pairs from the lower tail (less occlusion of face) give a better genuine distribution.

\section{Hairstyle-Balanced Accuracy Comparison}

How does accuracy of face recognition compare for females and males when the test data is balanced on hairstyle?
To approach this question, we first define what it means to ``balance on hairstyle".

We know that bald hairstyle is more frequent for males, and that mixed bald/not-bald) image pairs have different impostor and genuine distributions than not-bald/not-bald image pairs.
Therefore, to get a hairstyle-balanced comparison, we drop images with bald hairstyle.
Based on the numbers in Table~\ref{tab:count}, we can see that this reduces the number of images.

We know that facial hair is common for males and non-existent for females, and that mixed facial-hair/clean-shaven image pairs have different impostor and genuine distributions than clean-shaven/clean-shaven image pairs.
Therefore, to get a hairstyle-balanced comparison we also drop images with facial hair.
This step reduces the number of images by a larger fraction.

\input{latex_tables/res_table}

We also know that changes in the distribution of fraction of the image representing hair impact the impostor and genuine distribution.
Therefore, we want to balance the female and male image sets based on the portion of the 112x112 cropped face image that represents hair.
This is done by establishing a correspondence between female and male images based on the intersection-over-union (IoU) of the pixels in the hair regions of the images.
For each female image, select the male image with the highest IoU of the hair regions, and if this IoU is above a threshold of 0.8, the images are kept for the hairstyle-balanced accuracy evaluation.

The resulting hairstyle-balanced comparison of impostor and genuine distributions for females and males is shown in Figure~\ref{fig:exp_results}.
The change in the d-prime differences are tabulated in Table~\ref{tab:Results_table}.

The results in Figure~\ref{AAM_AAF_MORPH} are for a balanced subset from MORPH, with 2,127 images of African-American males(1024 subjects) and 2,127 images of African-American females(1564 subjects). The impostor and genuine distributions for the images balanced on hair dimensions show a fundamental change from the original dataset.
 
For ArcFace, the original d-prime score between male and female impostors is 0.509, whereas it is 0.129 after balancing hair dimensions. Thus, balancing hairstyles reduces $\approx$75\% in the false match rate gap between African males and females. A similar pattern holds for the gender-balanced matcher. The original d-prime score between male and female impostors is 0.410, whereas it is 0.085 after balancing hair dimensions. Thus, balancing hair dimensions reduces $\approx$79\% in the false match rate gap between African males and females.

\input{latex_tables/bootstrap}

The genuine distribution for males in the balanced subset is relatively unchanged but is slightly better for females in the balanced subset than the original dataset, reducing the gap in genuine distribution. In other words, the difference in false non-match rate between males and females significantly reduces after balancing for hair dimensions. For ArcFace, the original d-prime score between male and female genuine is 0.283, whereas it is 0.003 after balancing hair dimensions. Thus, balancing hair dimensions reduces $\approx$99\% in the false non-match rate gap between African-American males and females. Similar pattern holds for the gender-balanced matcher. The original d-prime score between male and female genuine is 0.375, whereas it is 0.042 after balancing hair dimensions. Thus, balancing hair dimensions reduces $\approx$89\% in the false non-match rate gap between African-American males and females.

After filtering for the hair dimensions, we present the results in Figure~\ref{CM_CF_MORPH} for a balanced subset from MORPH. with 684 images of Caucasian males (522 Subjects) and 684 images of Caucasian females (481 Subjects). The impostor and genuine distributions for the images balanced on hair dimensions show a fundamental change from the original dataset. Comparing to the original distributions, the gap in impostor distributions has reduced considerably. In other words, the difference in false match rate between males and females significantly reduces after hairstyle balancing. For ArcFace, the original d-prime score between male and female impostors is 0.246, whereas it is 0.187 after balancing hair dimensions. Thus, balancing hair dimensions reduces $\approx$24\% in the false-match rate gap between Caucasian males and females. A similar pattern holds for the gender-balanced matcher. The original d-prime score between male and female impostors is 0.287, whereas it is 0.113 after balancing hair dimensions. Thus, balancing hair dimensions reduces $\approx$61\% in the false match rate gap between African males and females. 

In addition, the genuine distribution seems to improve for both males and females after hairstyle balancing.
This, in turn, causes the false non-match rate for the hairstyle-balanced subset to be significantly lower than the original dataset. For Arcface, the original d-prime score between male and female impostors is 0.208, whereas it is 0.004 after balancing hair dimensions. Thus, balancing hair dimensions reduces $\approx$98\% in the false non-match rate gap between Caucasian males and females. Similar pattern holds for the gender-balanced matcher. The original d-prime score between male and female genuine is 0.260, whereas it is 0.028 after balancing hair dimensions. Thus, balancing hair dimensions reduces $\approx$89\% in the false non-match rate gap between African-American males and females.

Results for our own dataset are presented in Figure \ref{CM_CF_MFAD}. The balanced subset is 344 images of Caucasian males (178 Subjects), and 344 images of Caucasian females (149 Subjects). A similar patterns of shifts in impostor and genuine distribution is evident for our own dataset as for MORPH. For ArcFace, the original d-prime  between male and female impostors is 0.224, whereas it is 0.061 after hairstyle-balancing. Thus, hairstyle balancing accounts for $\approx$73\% of the false-match rate gap between Caucasian males and females. A similar pattern holds for the gender-balanced matcher. The original d-prime score between male and female impostors is 0.204, whereas it is 0.023 after hairstyle-balancing. Thus, hairstyle-balancing accounts for $\approx$89\% of the false match rate gap between males and females. 

The genuine distribution for males in the balanced subset seems to be relatively unchanged but is slightly better for females in the balanced subset than the original dataset, reducing the gap in genuine distribution. In other words, the difference in false non-match rate between males and females significantly reduces after balancing for hair dimensions. For ArcFace, the original d-prime score between male and female genuine is 0.228, whereas it is 0.000 after hairstyle-balancing. 

A similar pattern holds for the gender-balanced matcher. The original d-prime score between male and female genuine is 0.261, whereas it is 0.091 after hairstyle-balancing. Thus, hairstyle-balancing reduces $\approx$65\% in the false non-match rate gap between African-American males and females.

\section{Bootstrap Confidence Analysis}
The number of images in our hairstyle-balanced accuracy comparison is greatly reduced from the original dataset.
To analyze whether the results could possibly be due to a random sampling of that amount of data from the original dataset, we randomly selected 1000 times from the original dataset the number of subjects and images as in the final hairstyle-balanced subset. 
For both matchers, the d-prime for male-female impostors and genuine of hair dimensions balanced subset is not within one standard deviation of the mean of the randomly-sampled subsets, suggesting that these results are highly unlikely to be just by random chance.  
All the results are shown in Table~\ref{tab:bootstrap}.

\section{Conclusions and Discussion}

\paragraph{Cause of observed gender gap in accuracy.} 
One main contribution of this work is to document and explain a cause-and-effect understanding of the gender gap in face recognition accuracy.
The gender gap in accuracy that is initially observed with both ArcFace and with a gender-balanced matcher (as shown in Figure~\ref{fig:original_dist}) largely disappears when the test image set is balanced for females and males based on the amount of the image that represents the face (as shown in Figure~\ref{fig:exp_results}).
\vspace{-0.5em}
\paragraph{Quality of test data, not quantity of training data.}
One initial reaction to the observed gender gap in face recognition accuracy is that it must be caused by imbalance in the quantity of training %test 
data~\cite{Albiero_balance}.

Table \ref{tab:Results_table} contains a comparison of the d-prime differences between the ArcFace matcher, which was trained on the unbalanced MS1MV2 dataset, and a matcher resulting from explicitly balanced training data. For the original test data, the gender-balanced matcher had a smaller d-prime only for African-American impostor distributions.
For the hair-balanced test data, the gender-balanced matcher had smaller d' for the impostor distributions, but larger d' for the genuine distributions.
Thus the gender-balanced matcher made no consistent improvement toward more gender-balanced accuracy in the test data.
\vspace{-0.5em}
\paragraph{Accuracy of face attributes.}
Algorithms for classification of face attributes such as presence of beard or mustache is an active area of research ~\cite{Thom2020facial,Zheng2020survey}.
Our experiments  
suggests that there is still substantial room to improve the accuracy of such algorithms, especially for purposes such as detecting elements of facial hair.
It may also be useful to explore whether the accuracy of such algorithms varies across demographic groups.
\vspace{-0.5em}
\paragraph{In-the-wild, celebrity images.}
Our analysis is based on a controlled-scenario image set, rather than an in-the-wild, celebrity image set.
In-the-wild images will bring greater variation in pose, illumination, expression and occlusion.
Celebrity images will likely bring greater use of makeup and of photo-shopping of images.
Thus for some other datasets, analyzing the cause of an observed accuracy difference may involve characterizing levels of makeup use.

{\small
\bibliographystyle{ieee}
\bibliography{main.bib}
}

%-\vfill
%\clearpage
%\input{supplemental}

\end{document}

%% file: latex_figures/intro_fig.tex
\begin{figure}[tb]
    \centering
    \captionsetup[subfigure]{labelformat=empty}
    \begin{subfigure}[b]{0.775\linewidth}
        \centering
        \includegraphics[width=\linewidth]{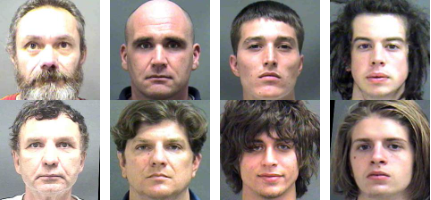}
        \caption{Male Pairs}
        \vspace{-0.5em}
    \end{subfigure}
    \hfill
    \begin{subfigure}[b]{0.18\linewidth}
        \centering
        \includegraphics[width=1\linewidth]{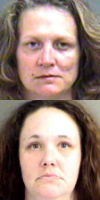}
        \caption{Female Pair}
        \vspace{-0.5em}
    \end{subfigure}
    \caption{Hairstyle balance of impostor image pairs. Each of the first four columns contains a male impostor pair.  The rightmost column contains a female impostor pair. Which of the male impostor pairs is best hairstyle-balanced to the female impostor pair? }
    \vspace{-1.5em}
    \label{fig:makeup_dist}
\end{figure}

%% file: latex_figures/original_dists.tex
\begin{figure*}[t]
  \begin{subfigure}[b]{1\linewidth}
    \centering
      \begin{subfigure}[b]{0.31\linewidth}
        \centering
          \includegraphics[width=1\linewidth]{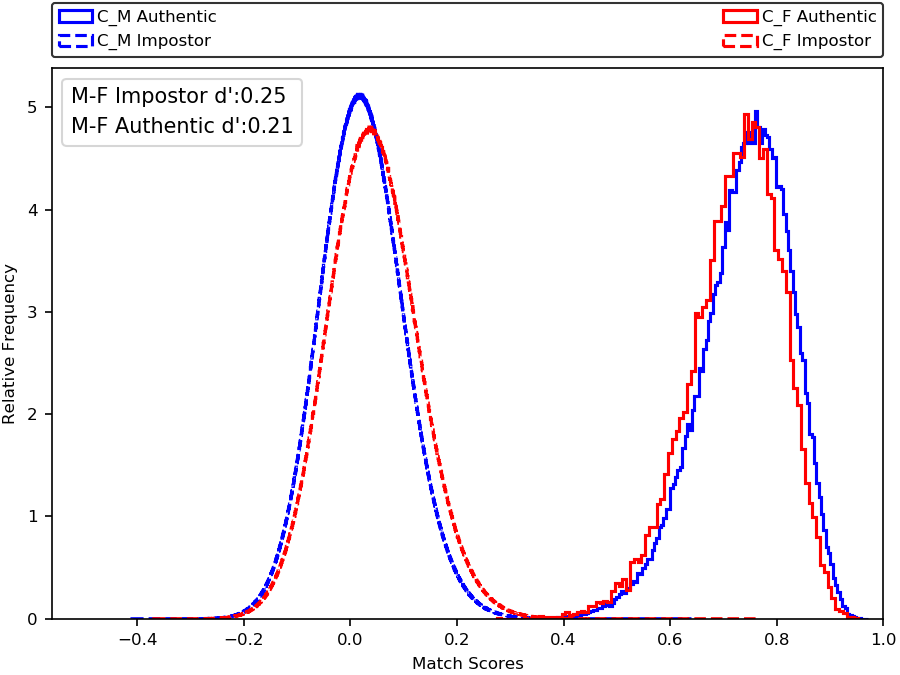}
      \end{subfigure}
      \hfill
      \begin{subfigure}[b]{0.31\linewidth}
        \centering
          \includegraphics[width=1\linewidth]{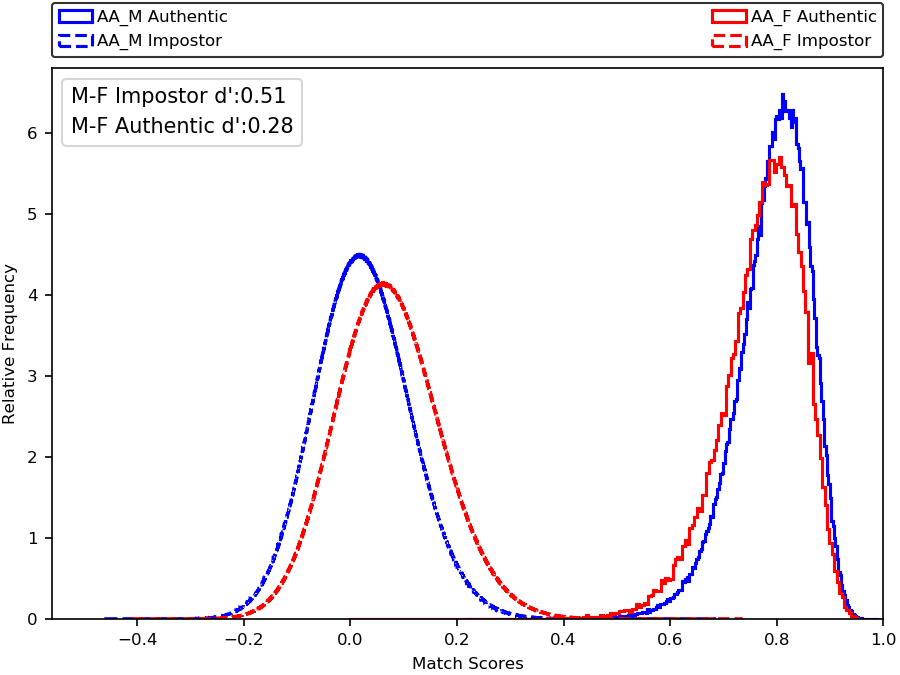}
      \end{subfigure}
      \hfill
      \begin{subfigure}[b]{0.31\linewidth}
        \centering
          \includegraphics[width=1\linewidth]{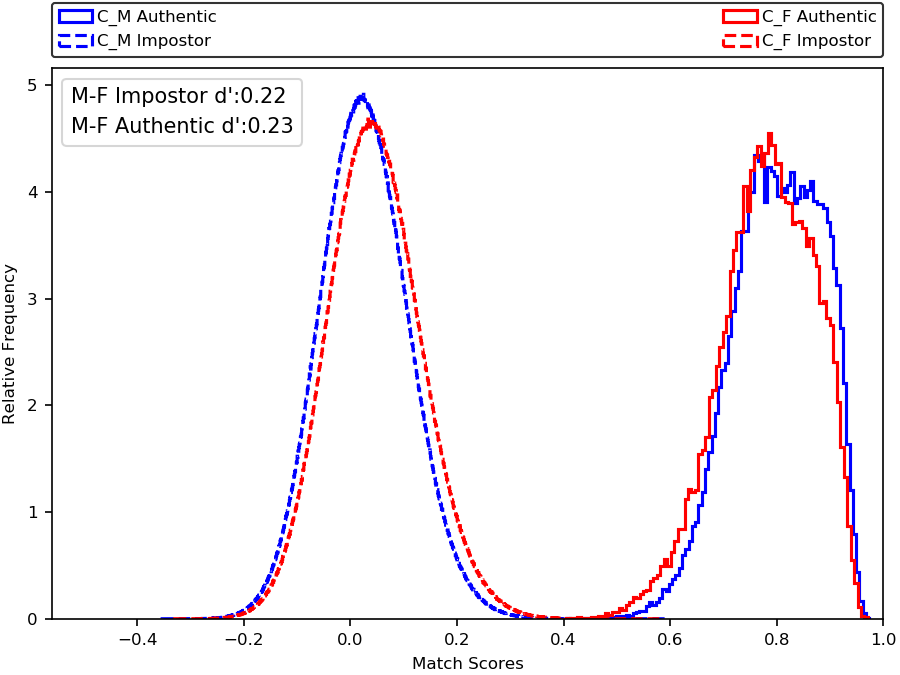}
      \end{subfigure}
    %   \vspace{1em}
  \end{subfigure}
%   \vspace{1em}
  \hfill
  \begin{subfigure}[b]{1\linewidth}
  \centering
      \begin{subfigure}[b]{0.31\linewidth}
        \centering
          \includegraphics[width=1\linewidth]{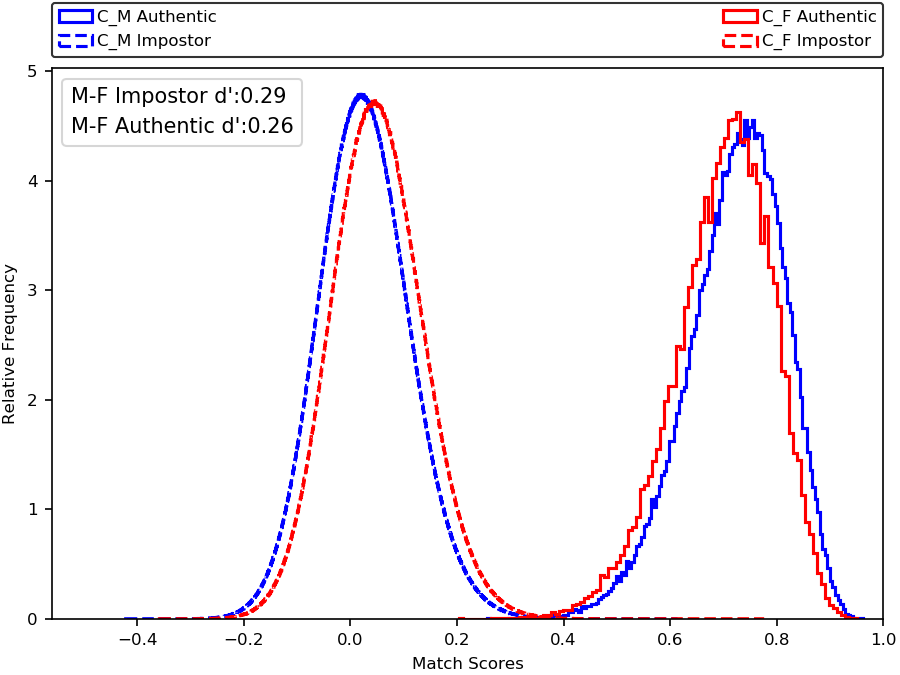}
            \caption{MORPH Caucasian}
            \vspace{-0.5em}
      \end{subfigure}
      \hfill
      \begin{subfigure}[b]{0.31\linewidth}
        \centering
          \includegraphics[width=1\linewidth]{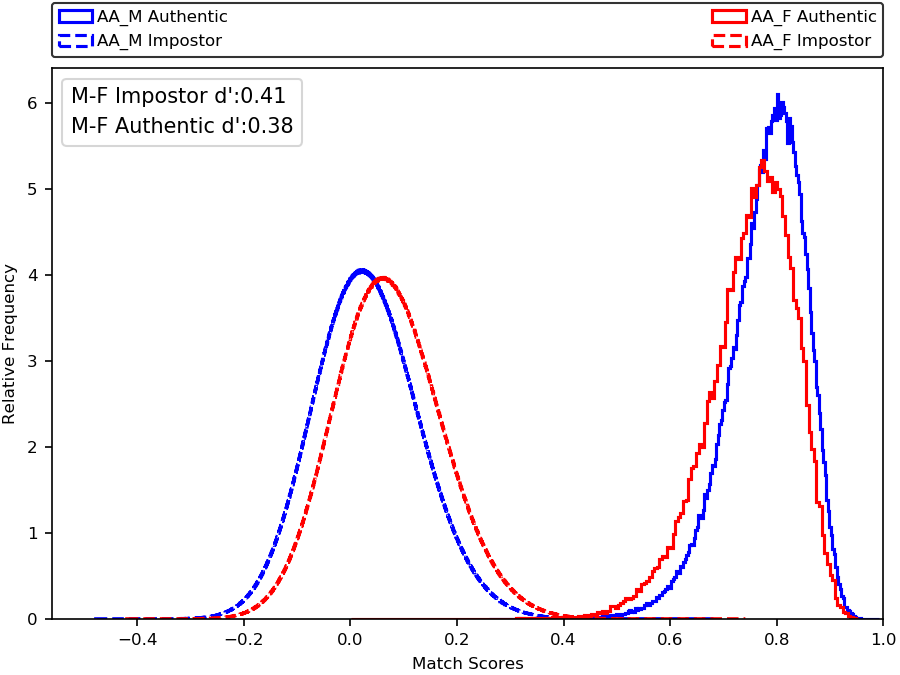}
         \caption{MORPH African-American}
         \vspace{-0.5em}
      \end{subfigure}
      \hfill
      \begin{subfigure}[b]{0.31\linewidth}
        \centering
          \includegraphics[width=1\linewidth]{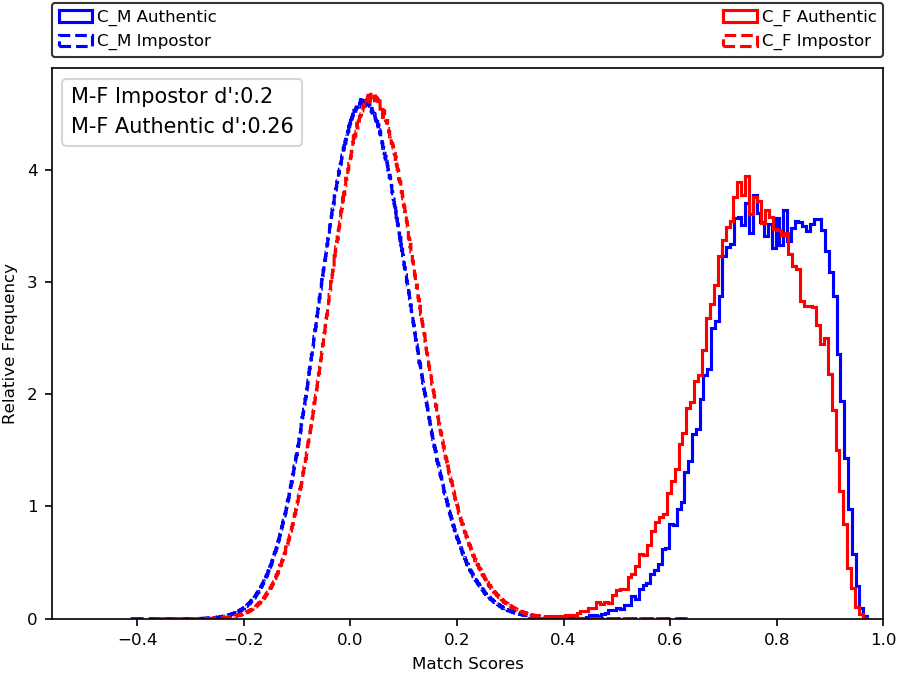}
            \caption{Our Own}
            \vspace{-0.5em}
      \end{subfigure}
  \end{subfigure}
  \caption{The ``gender gap'' in face recognition accuracy. The observation that females have worse impostor and genuine distribution was made by Klare et al.~\cite{Klare1} a decade ago. The same qualitative pattern is seen here. The top row compares impostor and genuine distributions for the MORPH and our own dataset using standard ArcFace matcher; bottom row shows comparisons for a version of ArcFace trained on a smaller, explicitly gender-balanced dataset.}
  \vspace{-1.5em}
  \label{fig:original_dist}
\end{figure*}

%% file: latex_tables/count.tex
\begin{table*}[]
\centering
\resizebox{\textwidth}{!}{%
\begin{tabular}{cccccc}
\multicolumn{6}{c}{\textbf{Facial Hair / No Facial Hair Classification Using MS Face + Rekognition Fusion}} \\ \hline
 \textbf{Dataset}& 
  \textbf{Prediction} &
  \textbf{Caucasian Males} &
  \textbf{Caucasian Females} &
  \textbf{African-American Males} &
  \textbf{African-American Females} \\ \hline
\multirow{2}{*}{\textbf{MORPH}} &
  \textbf{Facial Hair} &
  24,958(70\%) &
  2(0.1\%) &
  50,570(90\%) &
  132(0.5\%) \\
 &
  \textbf{No Facial Hair} &
  10,646(30\%) &
  10,938(99.9\%) &
  5,640(10\%) &
  24,721(99.5\%) \\ \hline
\multirow{2}{*}{\textbf{Our Own}} &
  \textbf{Facial Hair} &
   1710(25\%)&
   0 (0\%)&
  - &
  - \\
 &
  \textbf{No Facial Hair} &
   5293(75\%)&
   5444(100\%)&
  - &
  - \\\\
\multicolumn{6}{c}{\textbf{Bald/ Not-bald Classification Using BiSeNet Hair Ratio + MS Face}} \\ \hline
 \textbf{Dataset}&
  \textbf{Prediction} &
  \textbf{Caucasian Males} &
  \textbf{Caucasian Females} &
  \textbf{African-American Males} &
  \textbf{African-American Females} \\ \hline
\multirow{2}{*}{\textbf{MORPH}} &
  \textbf{Bald} &
  1,371(4\%) &
  3(0.1\%) &
  5,868(10.5\%) &
  30(0.2\%) \\
 &
  \textbf{Not Bald} &
  33,873(96\%) &
  10,937(99.9\%) &
  50,342(89.5\%) &
  24,823(99.8\%) \\ \hline
\multirow{2}{*}{\textbf{Our Own}} &
  \textbf{Bald} &
   30(0.4\%)&
   0(0\%)&
  - &
  - \\
 &
  \textbf{Not Bald} &
   6973(99.6\%)&
   5444(100\%)&
  - &
  -
\end{tabular}%
}
\caption{Facial hair and bald classification for Caucasians and African-Americans across gender. Females and males differ strongly in facial hair and baldness. 
% Face attribute algorithms can classify face images as having facial hair or not, and having a bald hairstyle or not.
% As expected, effectively no female face images show facial hair, whereas 30\% of Caucasian male and 10\% of African-American male facial images in Morph are classified as showing no facial hair. 75\% of Caucasian male images in Our Own are classified as showing no facial hair.
% Similarly, a bald hairstyle is classified for 0.4\% - 10\% of male face images, but is quite rare for females.
}
\vspace{-1.5em}
\label{tab:count}
\end{table*}

%% file: latex_figures/one_attribute.tex
\begin{figure*}[t]
    \begin{subfigure}[b]{1\linewidth}
        \begin{subfigure}[b]{0.32\linewidth}
          \includegraphics[width=\linewidth]{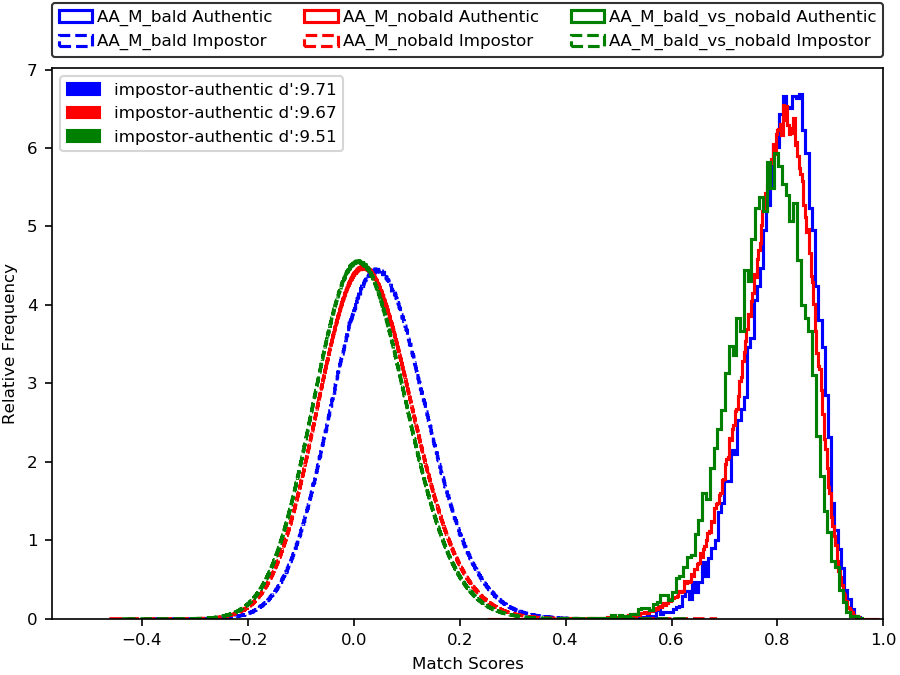}
          \caption{Impact of Bald Hairstyle On Accuracy}
          \vspace{-0.5em}
          \label{fig:sep_bald}
        \end{subfigure}
        \hfill %%
        \begin{subfigure}[b]{0.32\linewidth}
          \includegraphics[width=\linewidth]{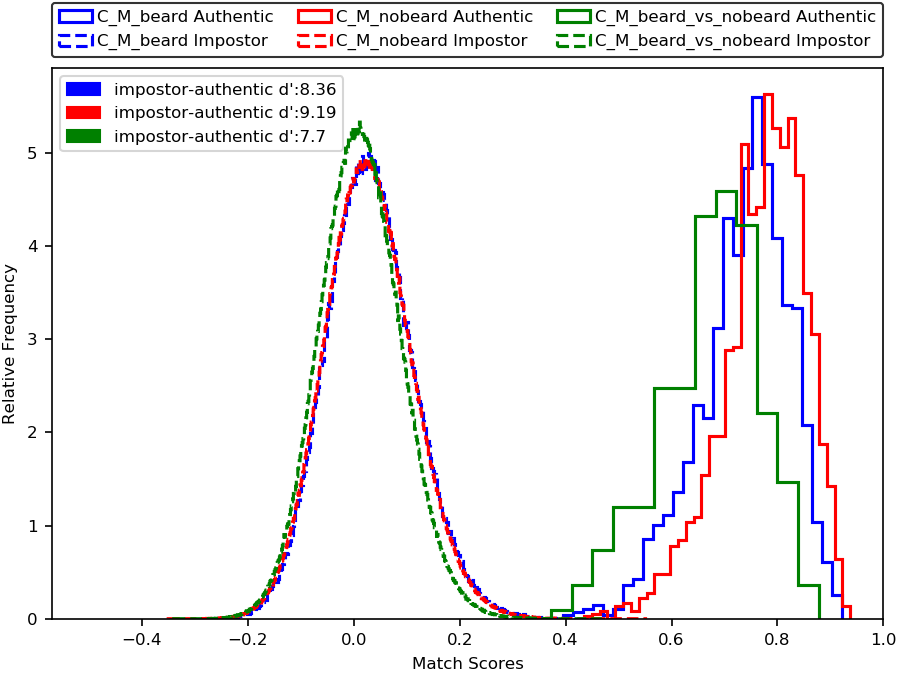}
          \caption{Impact of Facial Hair on Accuracy}
          \vspace{-0.5em}
          \label{fig:sep_beard}
        \end{subfigure}
        \hfill %%
        \begin{subfigure}[b]{0.32\linewidth}
          \includegraphics[width=\linewidth]{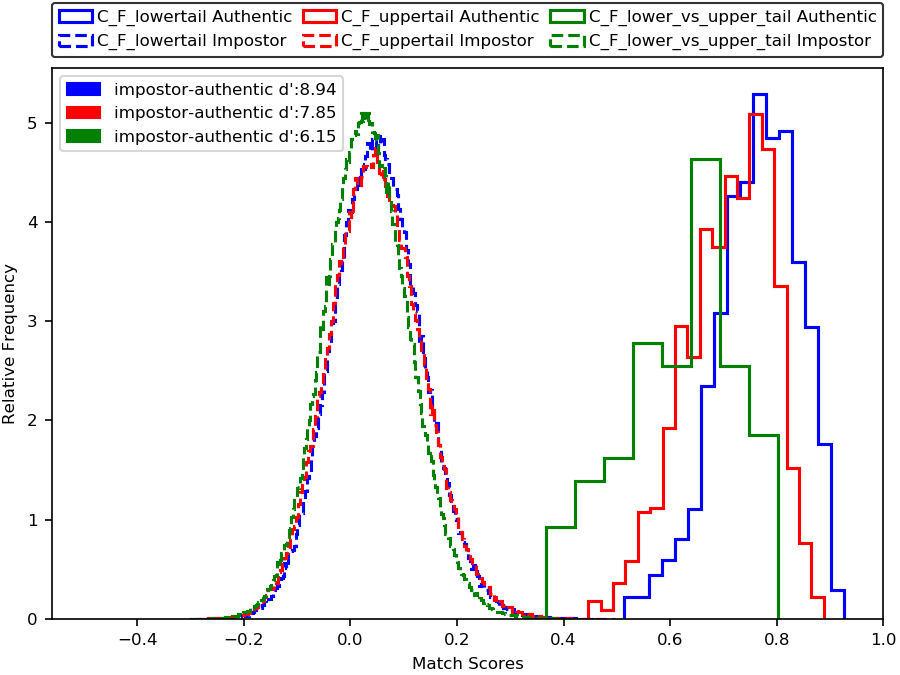}
          \caption{Impact of Fullness of Hairstyle}
          \vspace{-0.5em}
          \label{fig:sep_hair}
        \end{subfigure}
    \end{subfigure}
  \caption{Bald hairstyle, facial hair, and ``fullness'' of hairstyle impact the genuine and impostor distributions.}
  \vspace{-1em}
  \label{fig:sep_attribute}
\end{figure*}

%% file: latex_figures/hair_ratio_display.tex
\begin{figure}[t]
\centering
 \includegraphics[width=\linewidth]{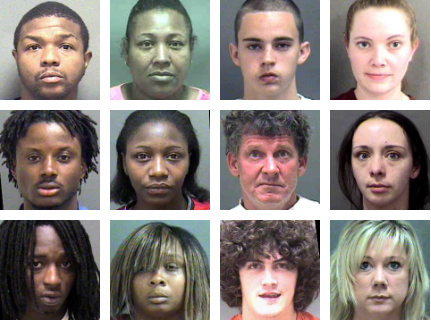}
 \caption{Increased hair ratio means greater face occlusion.
 Images in top row have $\approx$10\% of the 112x112 image occupied by pixels representing hair, middle row $\approx$30\%, and bottom row  hair $\approx$60\%.  Different distribution of hair ratio for a set of images can cause differences in the impostor and genuine distribution.}
 \vspace{-1.5em}
 \label{fig:hair_ratio_display}
\end{figure}

%% file: latex_figures/hair_ratio_dist.tex
\begin{figure}[t]
    \smallskip
    %\text{Your title}\par\medskip
    \begin{subfigure}[b]{1\linewidth}
      \begin{subfigure}[b]{0.32\linewidth}
        \centering
          \includegraphics[width=\linewidth]{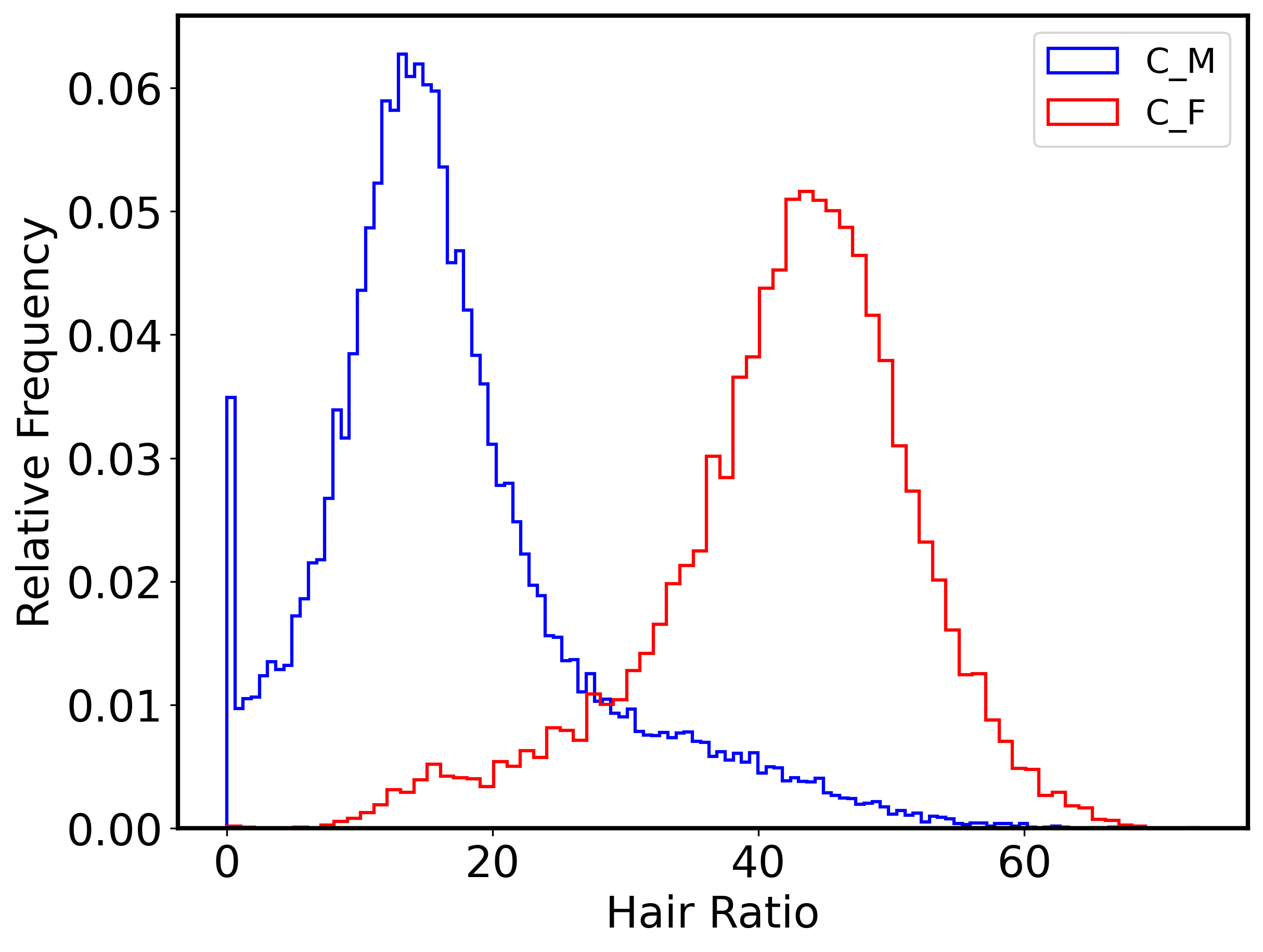}
          \caption{MORPH C}
          \vspace{-0.5em}
      \end{subfigure}
      \begin{subfigure}[b]{0.32\linewidth}
        \centering
          \includegraphics[width=\linewidth]{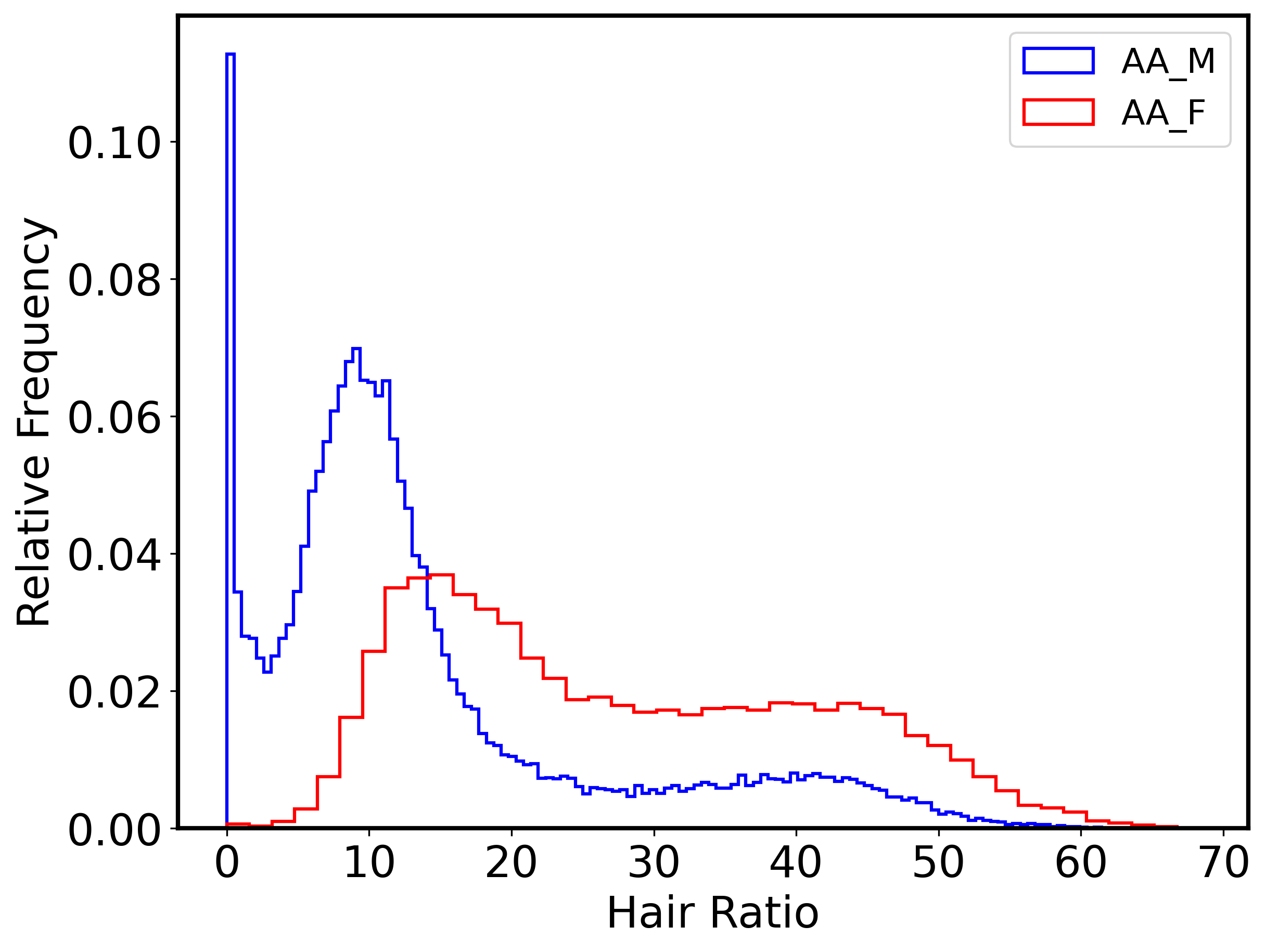}
          \caption{MORPH AA}
          \vspace{-0.5em}
      \end{subfigure}
      \begin{subfigure}[b]{0.32\linewidth}
        \centering
          \includegraphics[width=\linewidth]{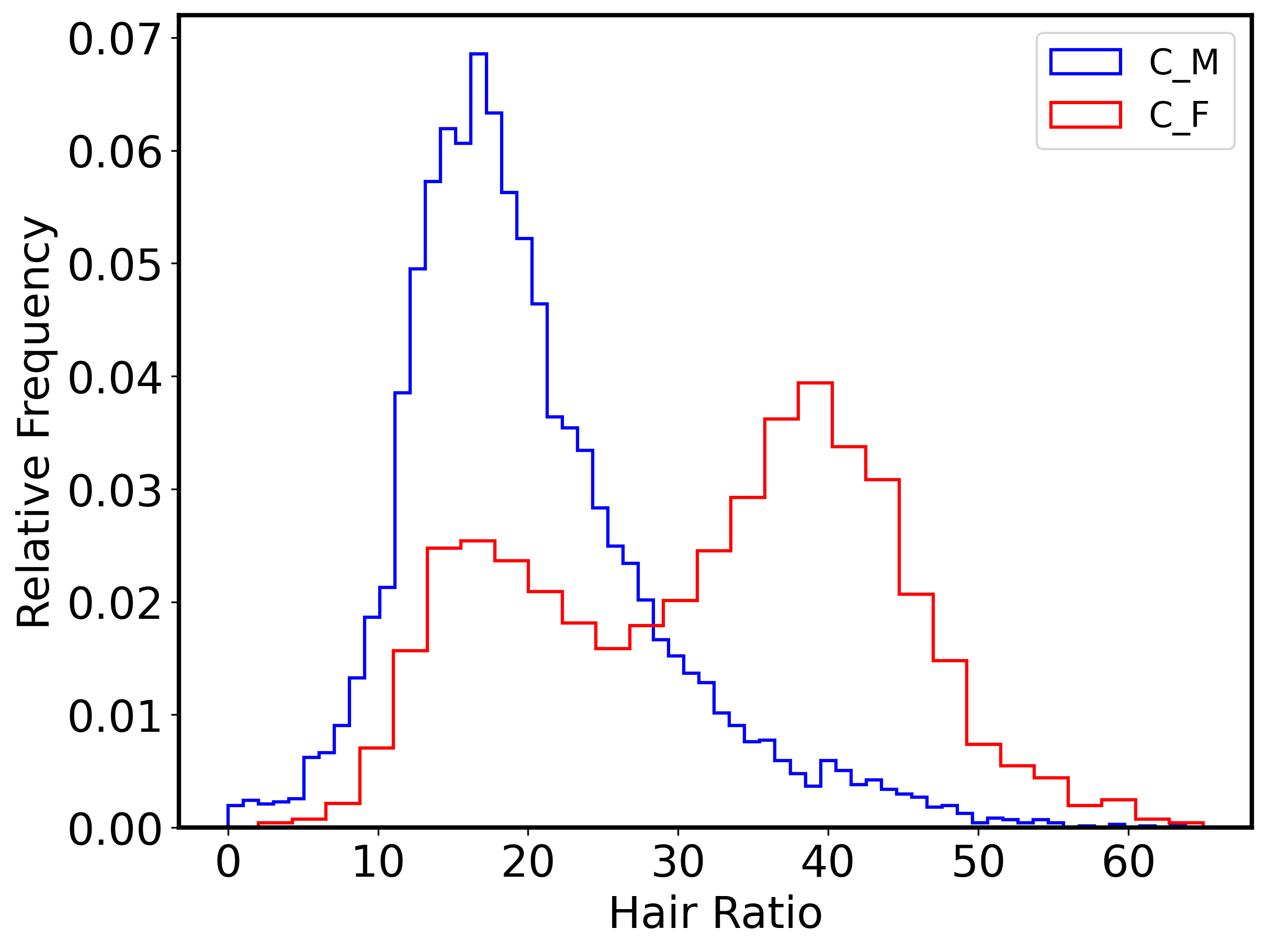}
          \caption{Our Own}
          \vspace{-0.5em}
      \end{subfigure}
    \end{subfigure}
    \caption{Distributions of fraction of cropped face image containing hair. Female images have, on average, a much larger of the face occluded by hair. 
    }
  \vspace{-1.5em}
  \label{fig:hair_ratio}
\end{figure}

%% file: latex_figures/exp_results.tex
\begin{figure*}[t]
    \begin{subfigure}[b]{1\linewidth}
        \centering
        \begin{subfigure}[b]{0.33\linewidth}
          \centering
          \includegraphics[width=1\linewidth]{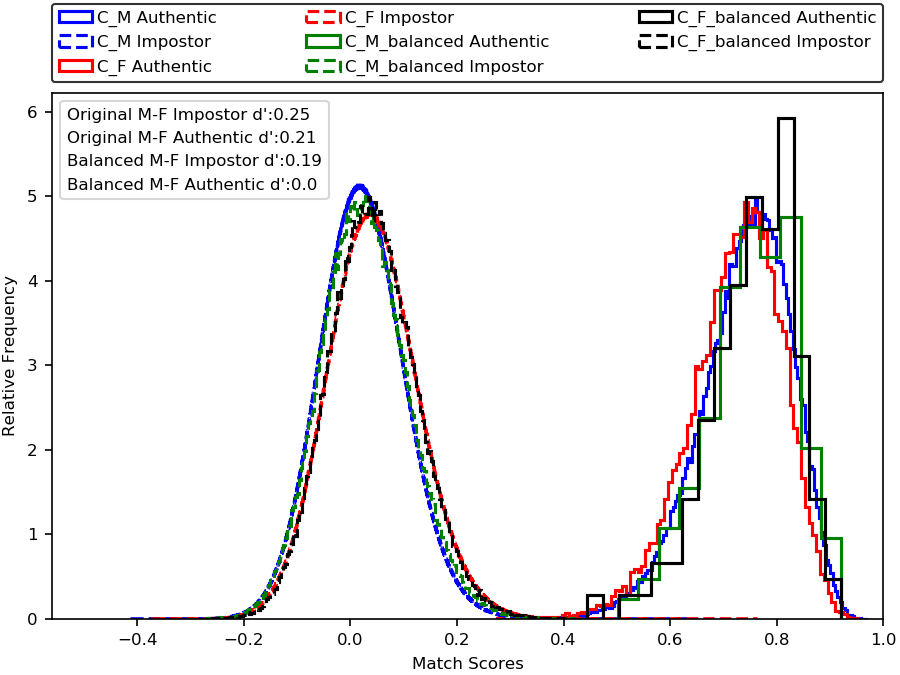}
        %   \vspace{1em}
        \end{subfigure}
        \begin{subfigure}[b]{0.33\linewidth}
          \centering
          \includegraphics[width=1\linewidth]{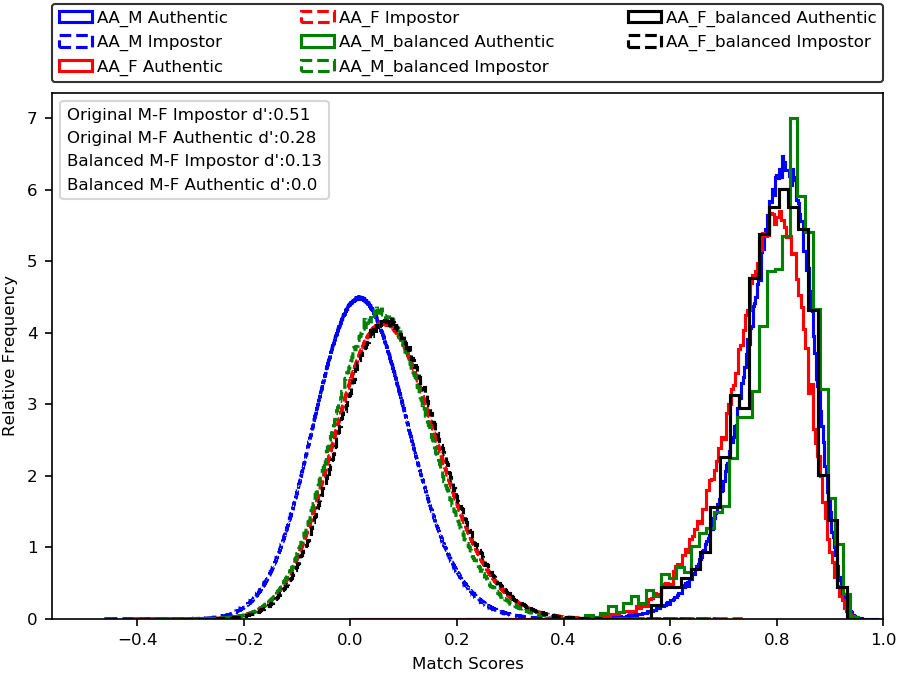}
        %   \vspace{1em}
        \end{subfigure}
        \begin{subfigure}[b]{0.33\linewidth}
          \centering
          \includegraphics[width=1\linewidth]{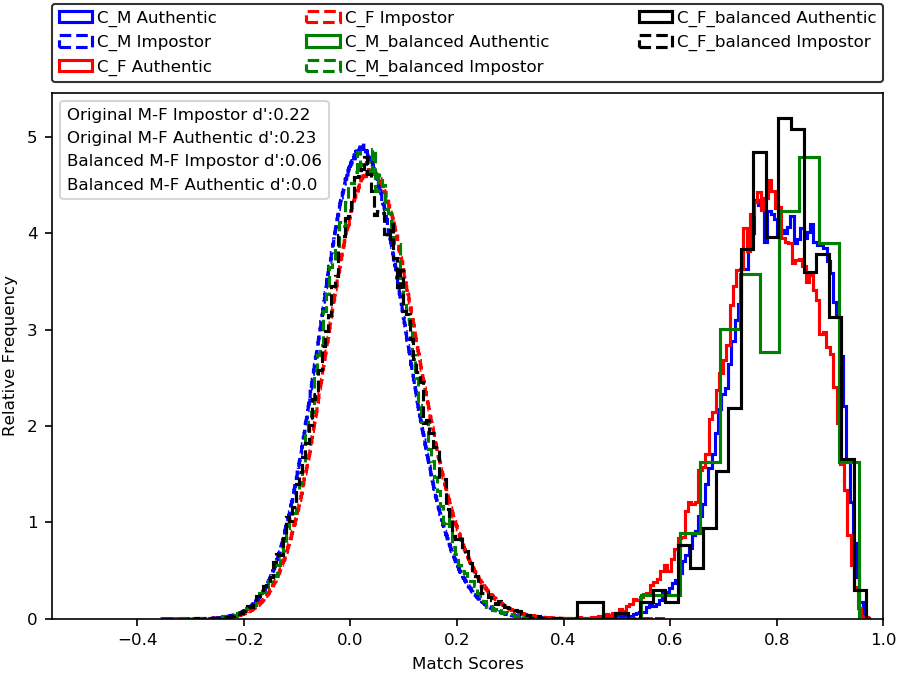}
        %   \vspace{1em}
        \end{subfigure}
    \end{subfigure}
    \begin{subfigure}[b]{1\linewidth}
        \centering
        \begin{subfigure}[b]{0.33\linewidth}
          \centering
          \includegraphics[width=1\linewidth]{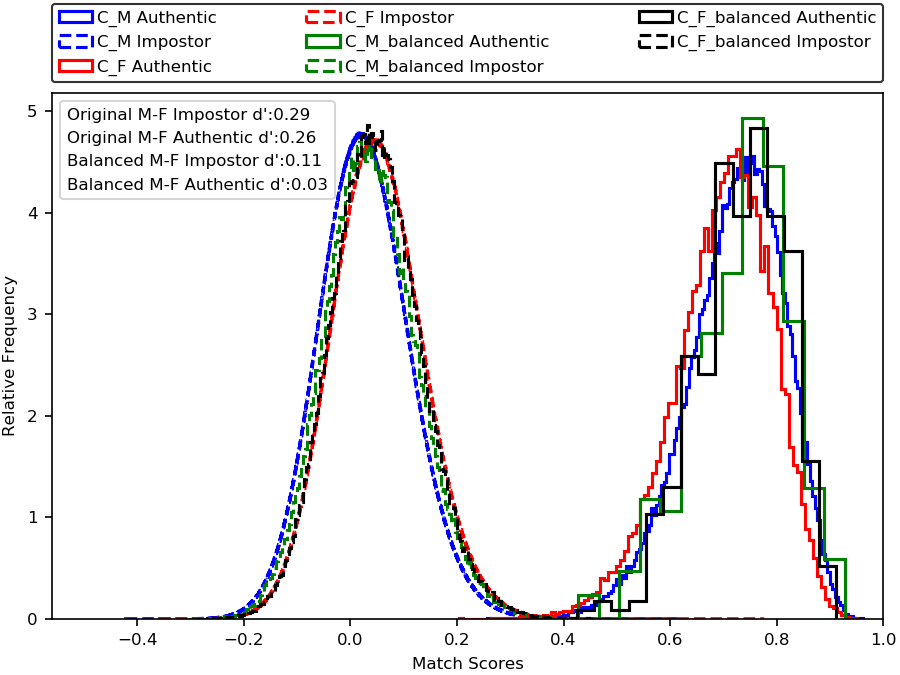}
          \caption{MORPH Caucasian}
          \vspace{-0.5em}
          \label{CM_CF_MORPH}
        \end{subfigure}
        \begin{subfigure}[b]{0.33\linewidth}
          \centering
          \includegraphics[width=1\linewidth]{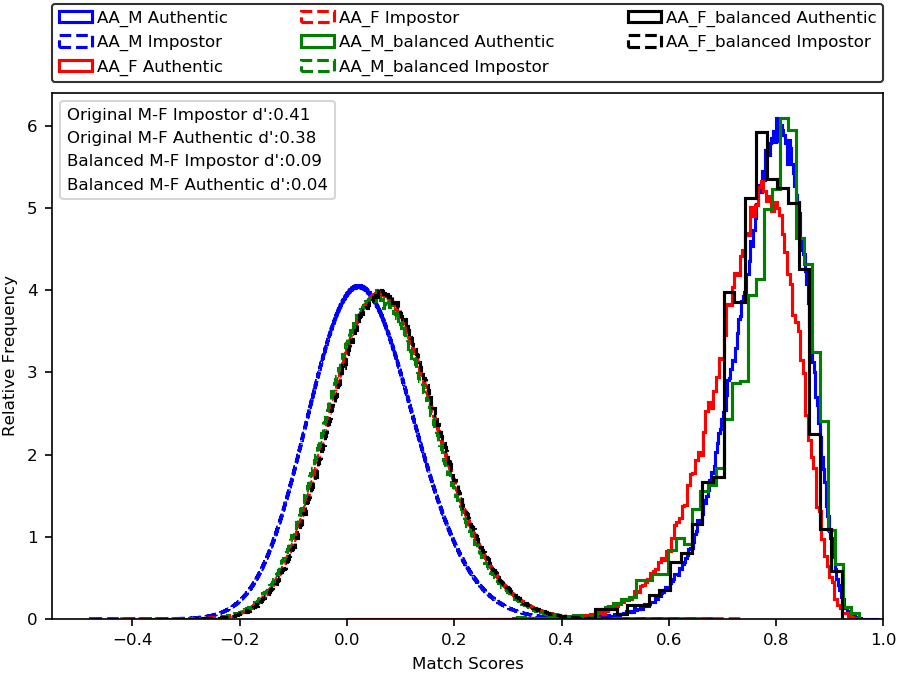}
          \caption{MORPH African-American}
          \vspace{-0.5em}
          \label{AAM_AAF_MORPH}
        \end{subfigure}
        \begin{subfigure}[b]{0.33\linewidth}
          \centering
          \includegraphics[width=1\linewidth]{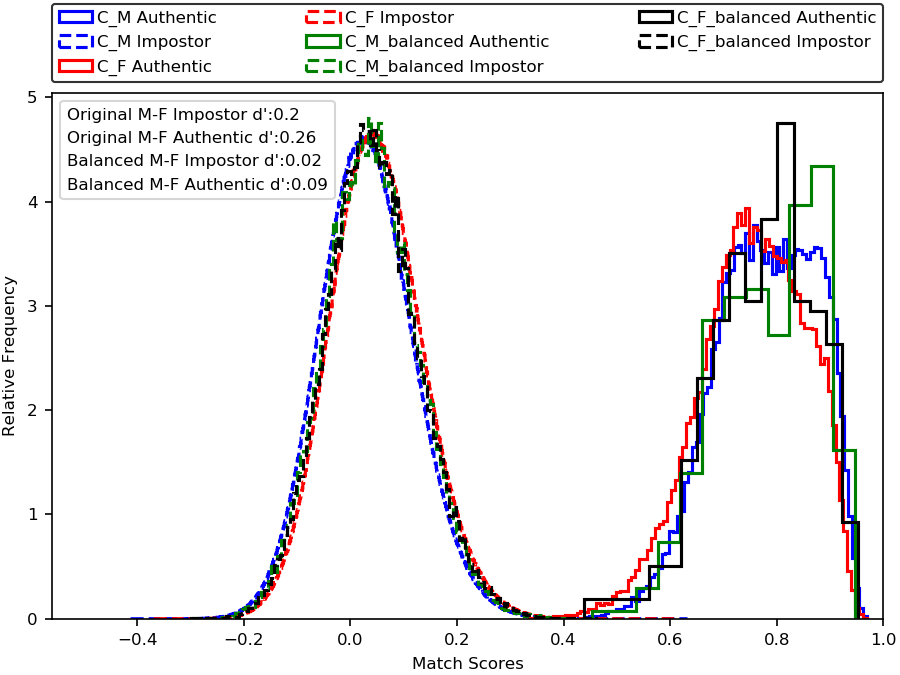}
          \caption{Our Own}
          \vspace{-0.5em}
          \label{CM_CF_MFAD}
        \end{subfigure}
    \end{subfigure}
  \caption{Impostor and genuine distributions for image sets that are
hairstyle-balanced across female / male. Top row results are for
ArcFace, bottom row results are for the gender-balanced matcher.}
\vspace{-1.5em}
  \label{fig:exp_results}
\end{figure*}

%% file: latex_tables/res_table.tex
\begin{table*}[]
\centering
\begin{tabular}{ccccccccc}
\textbf{} &
  \textbf{} &
  \textbf{} &
  \multicolumn{3}{c}{\textbf{Impostor}} &
  \multicolumn{3}{c}{\textbf{Genuine}} \\ \cline{4-9}
\textbf{Matcher} &
  \textbf{Dataset} &
  \textbf{Category} &
  \textbf{\begin{tabular}[c]{@{}c@{}}d-prime\\ before\end{tabular}} &
  \textbf{\begin{tabular}[c]{@{}c@{}}d-prime\\ after\end{tabular}} &
  \textbf{\begin{tabular}[c]{@{}c@{}}delta\\ d-prime\end{tabular}} &
  \textbf{\begin{tabular}[c]{@{}c@{}}d-prime\\ before\end{tabular}} &
  \textbf{\begin{tabular}[c]{@{}c@{}}d-prime\\ after\end{tabular}} &
  \textbf{\begin{tabular}[c]{@{}c@{}}delta\\ d-prime\end{tabular}} \\ \hline
\multirow{3}{*}{\textbf{ArcFace}} &
  \multirow{2}{*}{\textbf{MORPH}} &
  \textbf{C\_M vs C\_F} & 0.246 & 0.185 & -24\% & 0.208 & 0.004 & -98\% \\ & 
  &\textbf{AA\_M vs AA\_F} & 0.509 & 0.129 & -75\% & 0.283 & 0.003 & -99\% \\ \cline{2-9} 
 & \textbf{Our Own} &
  \textbf{C\_M vs C\_F} & 0.224 & 0.061 & -73\% & 0.228 & 0.000 & -100\% \\ \hline
\multirow{3}{*}{\textbf{Gender Balanced}} &
  \multirow{2}{*}{\textbf{MORPH}} &
  \textbf{C\_M vs C\_F} & 0.287 &0.113 &-61\% &0.260 &0.028 &-89\% \\&
   &
  \textbf{AA\_M vs AA\_F} &0.410 &0.085 &-79\% &0.375 &0.042 &-89\% \\ \cline{2-9}
 &
 \textbf{Our Own} &
  \textbf{C\_M vs C\_F} &0.204&0.023&-89\%&\color[HTML]{FD6864}0.261&\color[HTML]{FD6864}0.091&\color[HTML]{FD6864}-65\%
\end{tabular}%
\vspace{-0.5em}
\caption{d-prime for female / male impostor and genuine distributions. ``d-prime before" is for original test data, and ``d-prime after" is for hairstyle-balanced. Balancing on hairstyle decreases the gap between female and male impostor distributions and between female and male genuine distributions, for both African-Americans (AA) and Caucasians (C). The text in red is to show that the results might not be very reliable due to: (a) very few genuine pairs, and (b) shoulder in the high similarity tail of original genuine distribution.} 
\label{tab:Results_table}
\vspace{-1.5em}
\end{table*}

%% file: latex_tables/bootstrap.tex
% \begin{table*}[]
% \centering
% \small
% \resizebox{\textwidth}{!}{%
% \begin{tabular}{cccccccc}
% \textbf{} & \textbf{} & \multicolumn{3}{c}{\textbf{Impostor}} & \multicolumn{3}{c}{\textbf{Genuine}} \\ \cline{3-8} 
% \textbf{Matcher} & \textbf{Category} & \textbf{\begin{tabular}[c]{@{}c@{}}d-prime\\ balanced\end{tabular}} & \textbf{\begin{tabular}[c]{@{}c@{}}Mean d-prime\\ random\end{tabular}} & \textbf{\begin{tabular}[c]{@{}c@{}}Std.dev d-prime\\ random\end{tabular}} & \textbf{\begin{tabular}[c]{@{}c@{}}d-prime\\ balanced\end{tabular}} & \textbf{\begin{tabular}[c]{@{}c@{}}Mean d-prime\\ random\end{tabular}} & \textbf{\begin{tabular}[c]{@{}c@{}}Std.dev d-prime\\ random\end{tabular}} \\ \hline
% \multirow{2}{*}{\textbf{ArcFace}} & \textbf{C\_M vs C\_F} & 0.185 & 0.235 & 0.030 & 0.004 & 0.402 & 0.252 \\
%  & \textbf{AA\_M vs AA\_F} & 0.129 & 0.503 & 0.019 & 0.003 & 0.305 & 0.147 \\ \hline
% \multirow{2}{*}{\textbf{Gender Balanced}} & \textbf{C\_M vs C\_F} & 0.113 & 0.274 & 0.003 & 0.028 & 0.430 & 0.263 \\
%  & \textbf{AA\_M vs AA\_F} & 0.085 & 0.410 & 0.018 & 0.042 & 0.390 & 0.151
% \end{tabular}
% }
% \caption{Mean male-female impostor/genuine d-prime and std. dev. for 1000 random samples without replacement. ``d-prime balanced" is for hairstyle-balanced subset, ``Mean d-prime random" and ``Std.dev. d-prime random" are for 1000 random samples.  The hairstyle-balanced d-primes are not within one std. dev. of randomly sampled mean.}
% \label{tab:bootstrap}
% \end{table*}

% \usepackage{multirow}
% \usepackage{graphicx}
\begin{table*}[]
\centering
\resizebox{\textwidth}{!}{%
\begin{tabular}{ccccccccc}
\textbf{} &
   &
  \textbf{} &
  \multicolumn{3}{c}{\textbf{Impostor}} &
  \multicolumn{3}{c}{\textbf{Genuine}} \\ \cline{4-9}
\textbf{Matcher} &
  \textbf{Dataset} &
  \textbf{Category} &
  \textbf{\begin{tabular}[c]{@{}c@{}}d-prime\\ balanced\end{tabular}} &
  \textbf{\begin{tabular}[c]{@{}c@{}}Mean d-prime\\ random\end{tabular}} &
  \textbf{\begin{tabular}[c]{@{}c@{}}Std.dev d-prime\\ random\end{tabular}} &
  \textbf{\begin{tabular}[c]{@{}c@{}}d-prime\\ balanced\end{tabular}} &
  \textbf{\begin{tabular}[c]{@{}c@{}}Mean d-prime\\ random\end{tabular}} &
  \textbf{\begin{tabular}[c]{@{}c@{}}Std.dev d-prime\\ random\end{tabular}} \\ \hline
\multirow{3}{*}{\textbf{ArcFace}} &
  \multirow{2}{*}{\textbf{MORPH}} &
  \textbf{C\_M vs C\_F} &
  0.185 &
  0.235 &
  0.030 &
  0.004 &
  0.402 &
  0.252 \\
 &
   &
  \textbf{AA\_M vs AA\_F} &
  0.129 &
  0.503 &
  0.019 &
  0.003 &
  0.305 &
  0.147 \\ \cline{2-9}
 &
  \textbf{Our Own} &
  \textbf{C\_M vs C\_F} &
  0.061 &
  0.239 &
  0.091 &
  0.000 &
  0.276 &
  0.236 \\ \hline
\multirow{3}{*}{\textbf{Gender Balanced}} &
  \multirow{2}{*}{\textbf{MORPH}} &
  \textbf{C\_M vs C\_F} &
  0.113 &
  0.274 &
  0.003 &
  0.028 &
  0.430 &
  0.263 \\
 &
   &
  \textbf{AA\_M vs AA\_F} &
  0.085 &
  0.410 &
  0.018 &
  0.042 &
  0.390 &
  0.151 \\ \cline{2-9}
 &
  \textbf{Our Own} &
  \textbf{C\_M vs C\_F} &
  0.023 &
  0.214 &
  0.085 &
  \color[HTML]{FD6864}0.091 &
  \color[HTML]{FD6864}0.271 &
  \color[HTML]{FD6864}0.230
\end{tabular}%
}
\vspace{-0.25em}
\caption{Mean male-female impostor/genuine d-prime and std. dev. for 1000 random samples without replacement. ``d-prime balanced" is for hairstyle-balanced subset, ``Mean d-prime random" and ``Std.dev. d-prime random" are for 1000 random samples.  The hairstyle-balanced d-primes are not within one std. dev. of randomly sampled mean. The text in red is to show that balanced d-prime doesn't fall within one standard deviation. The most likely cause is mentioned in caption of Table \ref{tab:Results_table}.}
\label{tab:bootstrap}
\vspace{-1.5em}
\end{table*}